\pdfoutput=1

\documentclass[11pt]{article}

\usepackage[]{emnlp2021}

\usepackage{times}
\usepackage{latexsym}

\usepackage[T1]{fontenc}

\usepackage[utf8]{inputenc}

\usepackage{microtype}
\usepackage{graphicx} 
\usepackage{float} 
\usepackage{subfigure} 
\usepackage{wrapfig}
\usepackage{graphicx}
\usepackage{subfigure}
\usepackage{amsfonts}
\usepackage{bm}
\usepackage{array,amsmath}
\usepackage{amssymb}
\usepackage{dsfont}
\usepackage{float}
\usepackage{graphicx}
\usepackage{algorithm}
\usepackage{algorithmicx}
\usepackage{algpseudocode}
\usepackage{pgf,tikz}
\usepackage{mathrsfs}
\usepackage{multirow}
\usepackage{booktabs}
\usetikzlibrary{arrows}
\usepackage{threeparttable}

\def\figref#1{Figure~\ref{fig:#1}}
\def\figlabel#1{\label{fig:#1}\label{p:#1}}

\def\tabref#1{Table~\ref{tab:#1}}
\def\tablabel#1{\label{tab:#1}\label{p:#1}}

\def\eqref#1{Eq.~\ref{eqn:#1}}

\newcounter{notecounter}
\newcommand{\enotesoff}{\long\gdef\enote##1##2{}}
\newcommand{\enoteson}{\long\gdef\enote##1##2{{
			\stepcounter{notecounter}
			{\large\bf
				\hspace{1cm}\arabic{notecounter} $<<<$ ##1: ##2
				$>>>$\hspace{1cm}}}}}
\enoteson
\enotesoff
\long\def\eat#1{}

\title{Locating Language-Specific Information in Contextualized Embeddings}

\author{Sheng Liang, Philipp Dufter, Hinrich Sch\"utze\\
	Center for Information and Language Processing (CIS) \\
	LMU Munich, Germany\\
	{\tt \{shengliang, philipp\}@cis.lmu.de}}

\begin{document}
\maketitle
\begin{abstract}
Multilingual pretrained language models (MPLMs) 
exhibit multilinguality and are well
suited for transfer across languages. Most MPLMs
are trained in an unsupervised fashion and the relationship
between their objective and multilinguality is
unclear. More specifically, the question whether MPLM
representations are language-agnostic or they simply
interleave well with learned task prediction heads
arises. In this work, we locate language-specific information
in MPLMs and identify its dimensionality and the layers where this information occurs. We show that language-specific information is scattered across many dimensions, which can be projected into a linear subspace. 
Our study contributes to a better understanding of  MPLM
representations, going beyond treating them as unanalyzable
blobs of information.
\end{abstract}

\section{Introduction}

Multilingual contextualized language models (e.g., \citealp{devlin-etal-2019-bert,conneau-etal-2020-unsupervised}) have been shown to exhibit a great degree of multilinguality as for example measured by zero-shot crosslingual transfer \cite{hu2020xtreme}. Having multilingual models is useful as it is easier to deploy and maintain a single multilingual model rather than many monolingual ones. Further, low resource languages can benefit from transfer and less annotated data might be required. However, it is an active research question how current multilingual models work \cite{pires-etal-2019-multilingual,wu-dredze-2019-beto} and whether the models are truly language agnostic \cite{gonen2020s,libovicky-etal-2020-language,zhao2020inducing,choenni2020does}. 

Most prior work such as \cite{libovicky-etal-2020-language,zhao2020inducing} aims at creating language agnostic representations by subtracting the mean of language specific representations. \citet{gonen2020s} tries to extract word level translations from mBERT \cite{devlin-etal-2019-bert}.

We continue this line of research to decode interpretable language-specific information from multilingual representations, by analyzing the dimensionality of language-specific subspaces and how language-specific information evolves across different intermediate layers. To investigate this, we compare two linear projection methods, DensRay \cite{dufter-schutze-2019-analytical} and Linear Discriminant Analysis (LDA) \cite{fisher1936use}. Our experiments with three probing tasks, Language Identification, Linguistic Typology and Language Similarity, show that language-specific information is scattered across many dimensions. The dimensionality of the subspace is determined by the tasks. However, with our proposed methods there exists an upper bound on the dimensionality which is approximately equal to the number of languages in the model. We also find that languages are better separated in lower layers. 

With this knowledge, we propose an activity regularization approach in Appendix~\ref{appendix:table}, which potentially encourages the model to preserve language-specific information during finetuning. We apply this approach when finetuning mBERT on NER and POS tagging tasks and show that it achieves small improvements on zero-shot cross-lingual transfer.\footnote{Our code for the experiments is available on
\\	
\url{https://github.com/liangsheng02/Locating-Language-Information}.}

\eat{
Also, previous work focused on zero-shot transfer \cite{pires-etal-2019-multilingual,wu-dredze-2019-beto,hu2020xtreme} demonstrated that multilingual contextualized language models performed remarkable cross-lingual transfer capabilities on many tasks. However it's still unexplained that how language-specific information is stored in the models, and how its working mechanisms when transfer across languages.
}

\section{Related Work}
Much recent work has examined how language information is stored in multilingual representations. \newcite{pires-etal-2019-multilingual} construct a tree structure to describe language similarities via hierarchical clustering and CCA similarity scores. \newcite{libovicky-etal-2020-language} assume that mBERT’s representations have a language-neutral component and
a language-specific component. They remove the
language-specific component by subtracting the language
centroids. In a similar vein, \newcite{zhao2020inducing}
induce language agnostic representations and explore batch
normalization to decrease the distance between languages for
better cross-lingual transfer, achieving improvement on XNLI and RFEval. Further, \newcite{gonen2020s} extract an empirical language-identity subspace and language-neutral subspace in mBERT; to achieve this, they present a linear projection method based on null-space transformation yielded by iterative classification. This work mainly focuses on word translation and the dimensionality of the subspaces is not well studied. Similarly, \newcite{choenni2020does} investigate the language agnosticity using typological probing tasks. \newcite{huang2021disentangling} use a learning approach to achieve a linear syntax-subspace in mBERT, in which syntactic information is shared across languages. 
Researching a different aspect, \newcite{conneau-etal-2020-emerging} show that sharing upper layers is essential for obtaining multilingual models. Related to this, \newcite{muller-etal-2021-first} show that the lower part of mBERT operates as a multilingual encoder which is critical for cross-lingual transfer, and a language-agnostic predictor which is less important in cross-lingual transfer. 

Overall, we follow this line of research, but we set the focus on locating the language-specific information both in terms of \emph{dimensionality} of the subspace and in which intermediate \emph{layers} this information is contained. In addition, we propose an \emph{activity regularization} approach based on our findings. 

\section{Methods}

\subsection{Locating Language-Specific Information}

We aim at locating language-specific linear subspaces. Therefore, we compare two linear projection methods, DensRay \cite{dufter-schutze-2019-analytical} and LDA \cite{fisher1936use}. DensRay learns an orthogonal transformation of an embedding space to find a subspace of certain linguistic features. It maximizes the distances of embeddings between different languages and minimizes them within each language. Similarly, LDA identifies directions in the space that separate classes (languages).

The methods require representations from different
classes. More specifically, we obtain vectors
$(x^{l_1}_i)_{i \in 1,2,\dots}
(x^{l_2}_i)_{i \in 1,2,\dots}\in \mathbb{R}^d$, \ldots for languages
$l_1,l_2,\dots$. When we feed these vectors together with
the language information to the algorithms we obtain
projection matrices $Q \in \mathbb{R}^{d \times d}$ where the n-th dimension contains
the n-th most language information.
Please consult the references
for more details on how the
matrices are computed.

Both methods are computable with analytical
solutions. The upper bound of the subspace dimensionality is
characterized by the matrix rank in their objective function
(see \ref{appendix:upper}). There are two main differences
between DensRay and LDA: 1) DensRay yields an orthogonal
transformation matrix, while orthogonality is not guaranteed
for LDA  \cite{ye2006computational}. 2) DensRay only
considers the centroids of each language, while LDA
additionally utilizes the centroid of the entire space. We choose both methods as LDA is a canonical choice  and DensRay is an analytical version of Densifier \cite{rothe-etal-2016-ultradense}, which is an established method in natural language processing.

\section{Experiments}

\begin{figure}[t]
	\centering
	\def\wm{0.9\linewidth}
		\begin{minipage}[t]{1\linewidth}
			\centering
			\includegraphics[width=\wm]{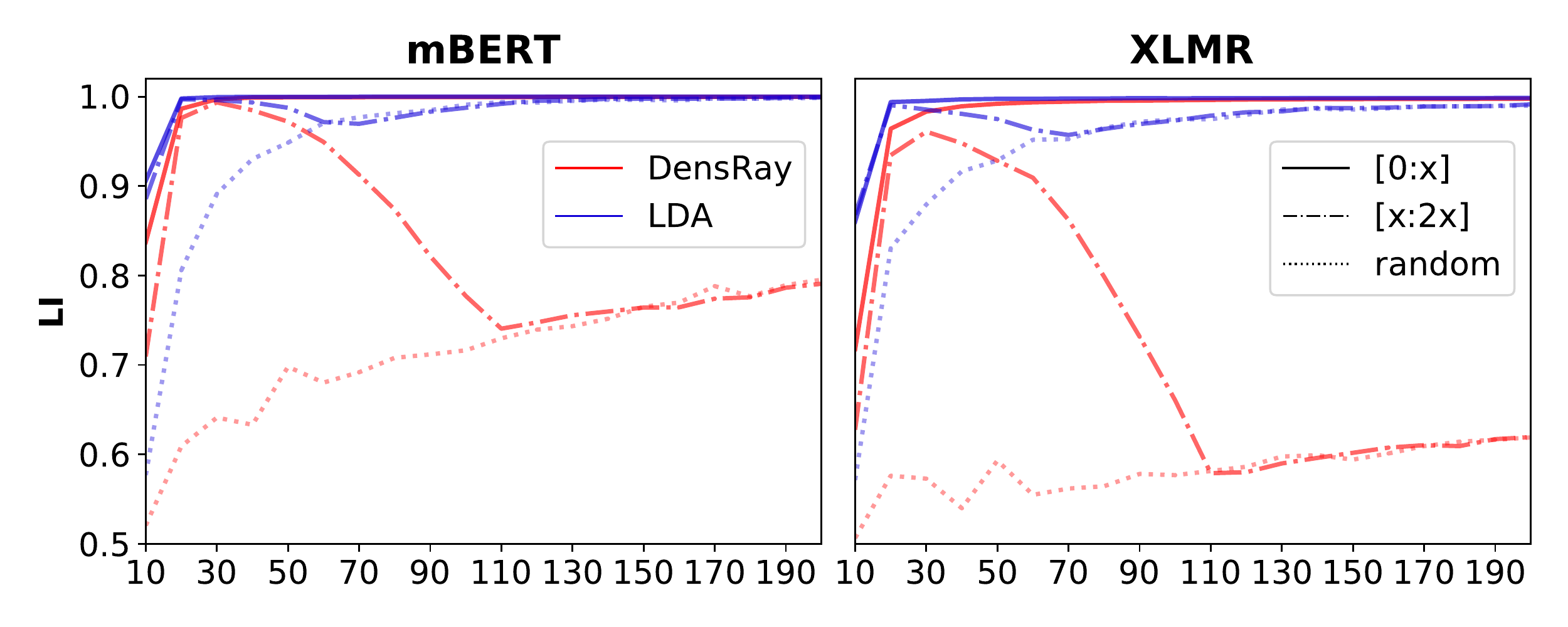}\\
			\vspace{-7pt}
			\includegraphics[width=\wm]{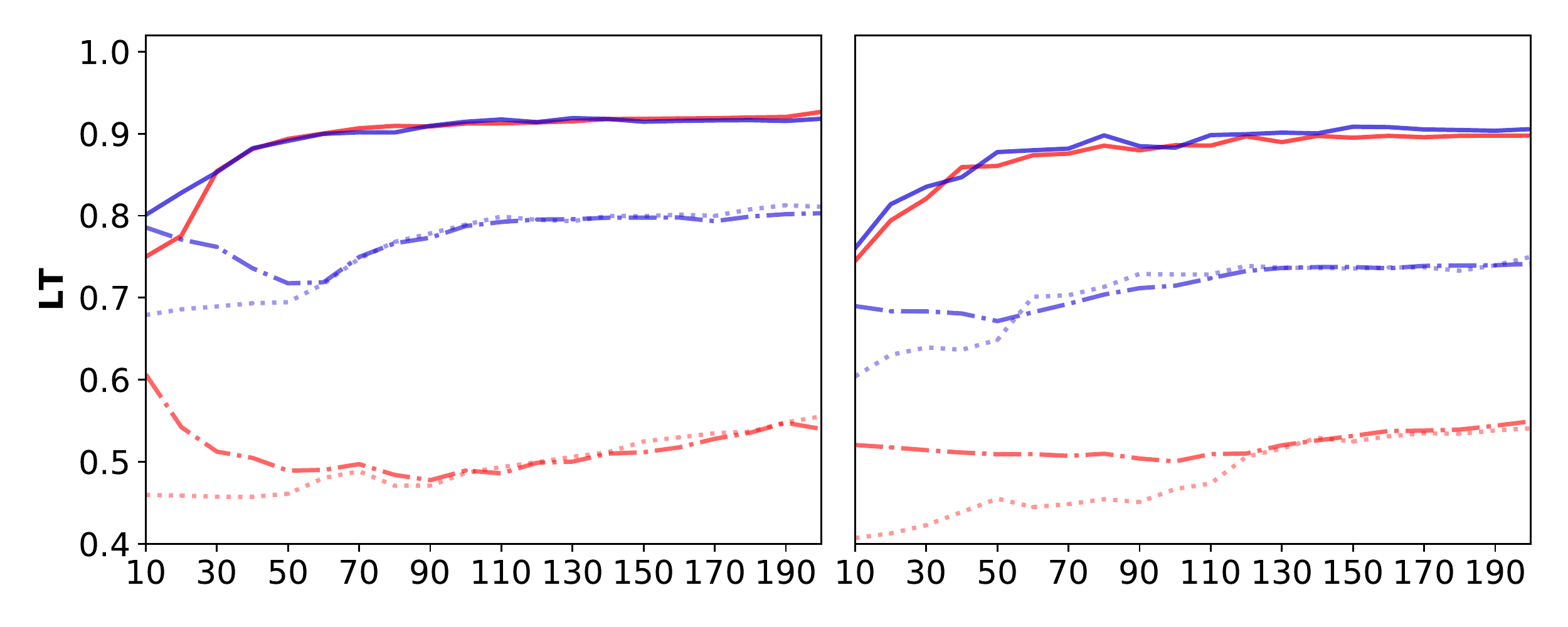}\\
			\vspace{-7pt}
			\includegraphics[width=\wm]{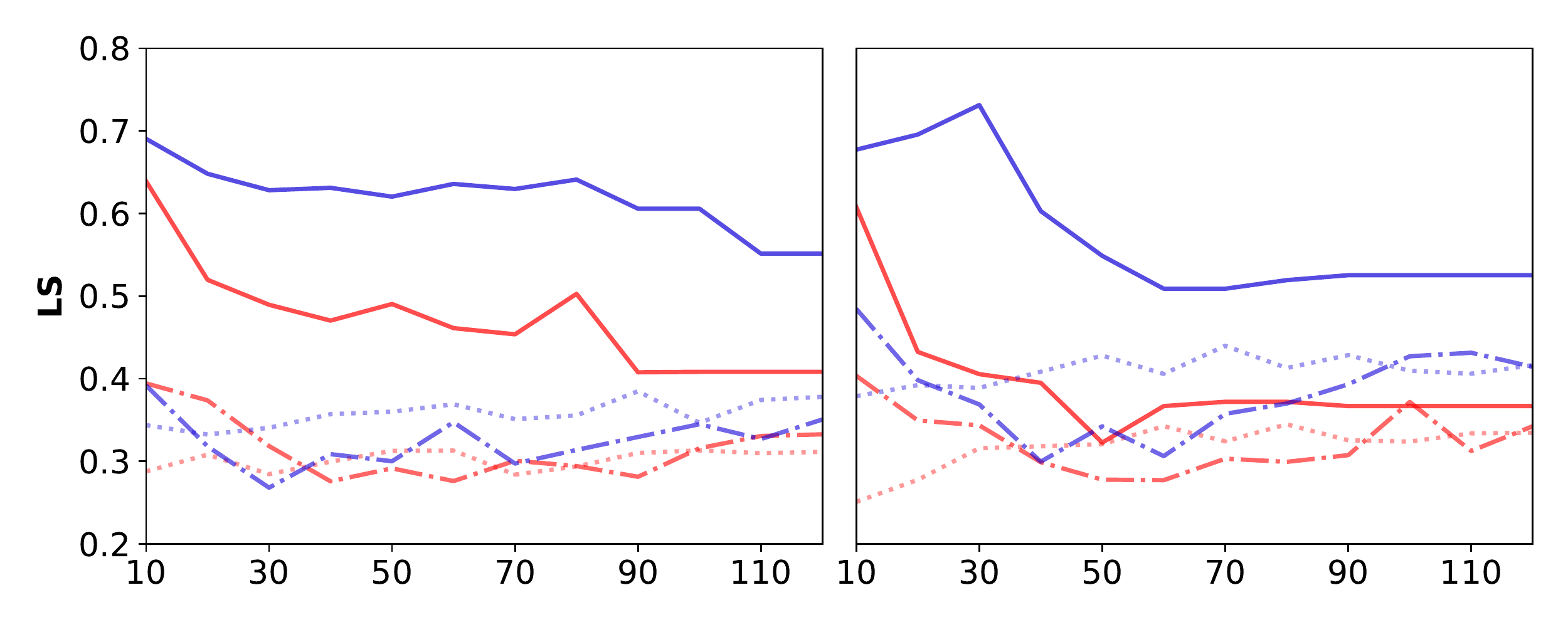}
		\end{minipage}%
	\caption{Probing tasks LI, LT and LS with
          $x$-dimensional subspaces on the 12th layer. We
          compare: the first $x$ dimensions [0:$x$], the
          next $x$ dimensions [$x$:2$x$] and $x$
          random dimensions.}
	\figlabel{subspace}
\end{figure}

\begin{figure}[t]
	\centering
		\def\wm{0.9\linewidth}
		\begin{minipage}[t]{1\linewidth}
			\centering
			\includegraphics[width=\wm]{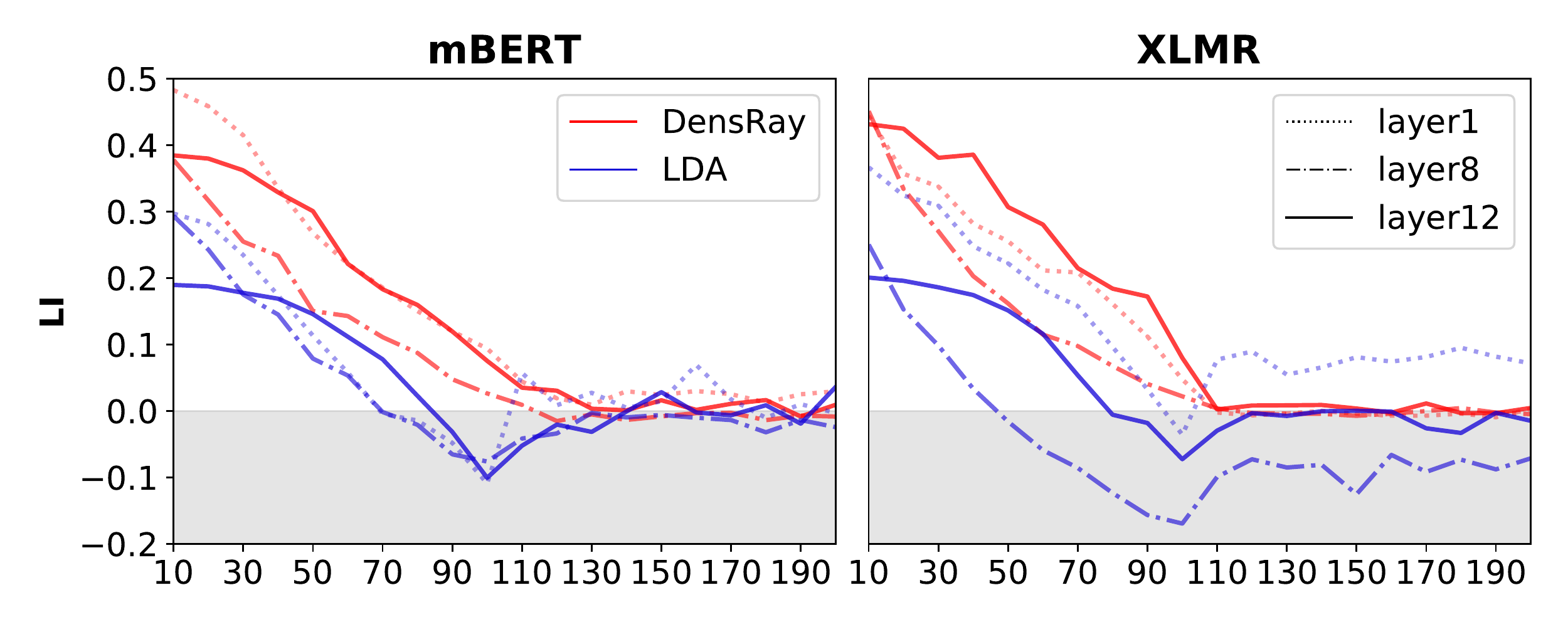}\\
			\vspace{-7pt}
			\includegraphics[width=\wm]{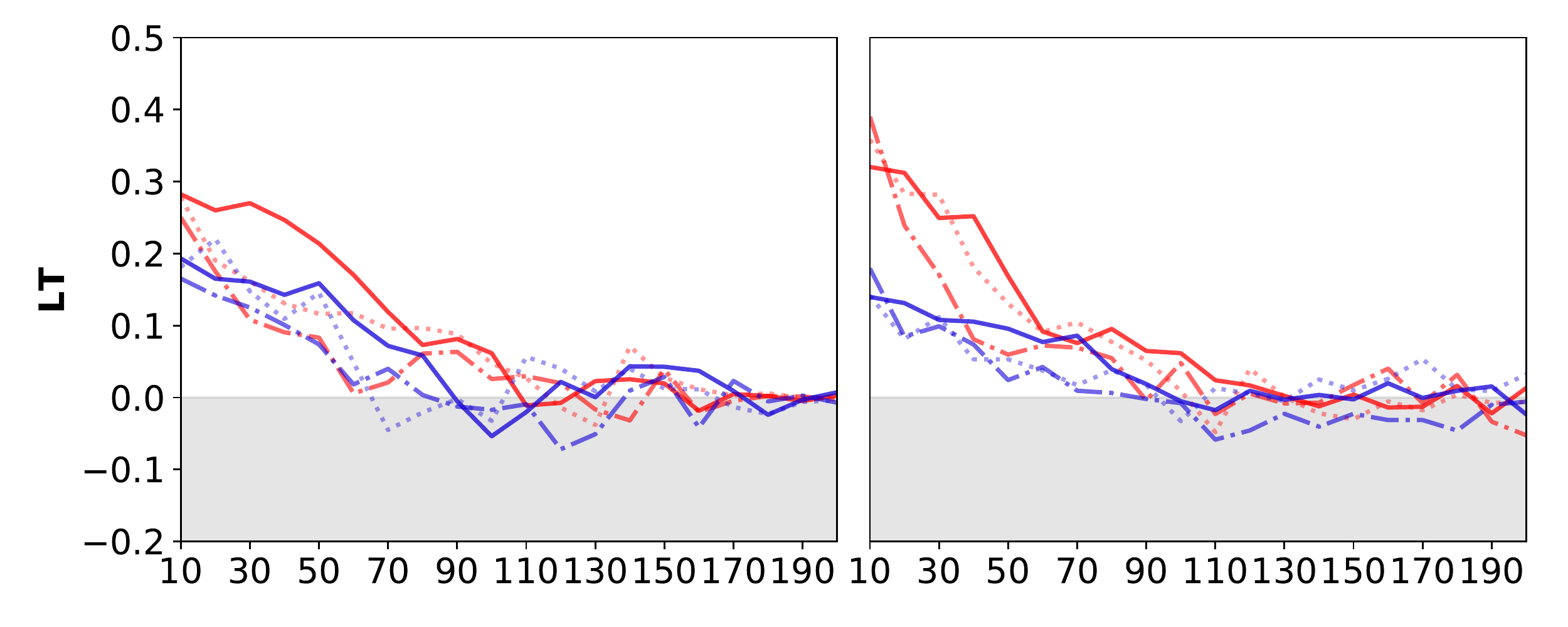}\\
			\vspace{-7pt}
			\includegraphics[width=\wm]{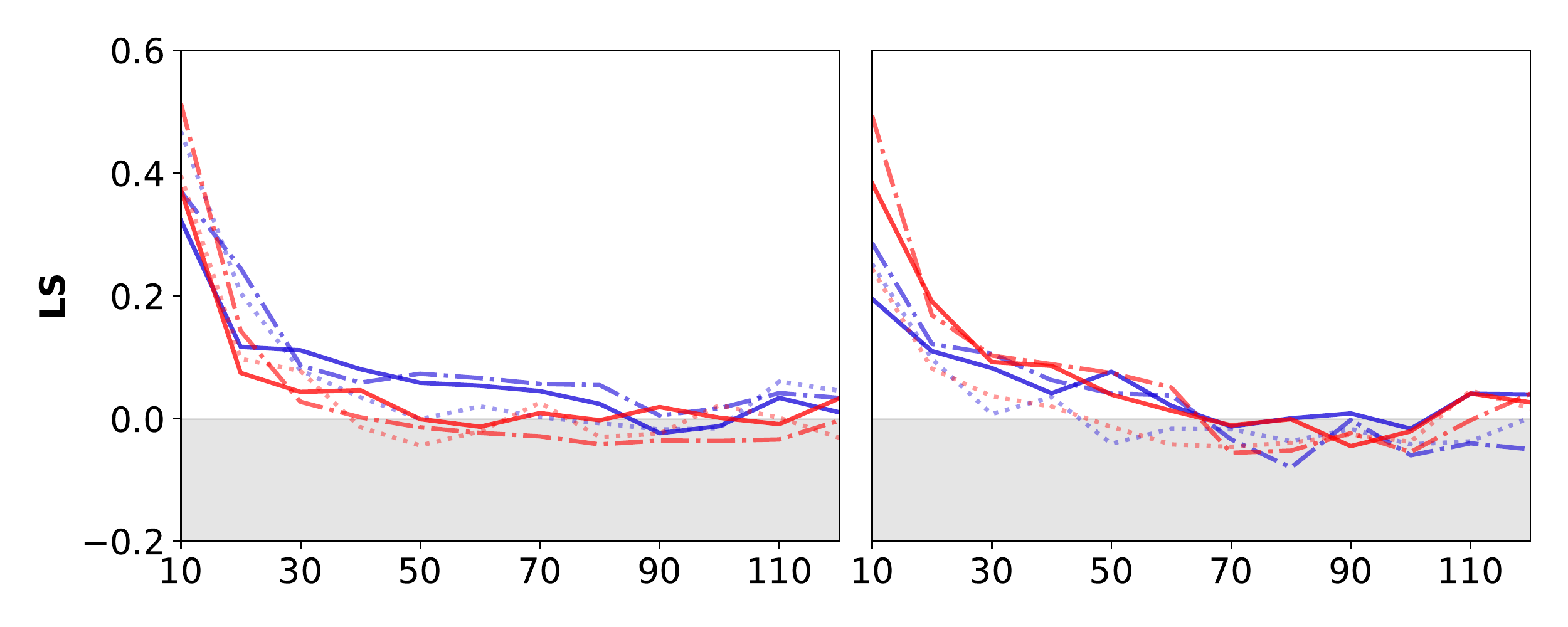}
		\end{minipage}%
	\caption{Probing tasks LI, LT and LS when
          considering 10 dimensions. For 50 on the x-axis,
          we report the difference between the accuracy when
          using dimensions 41--50
vs.\ 10 random dimensions.}
	\figlabel{diff}
\end{figure}

\subsection{Computing Language Specific Subspaces}
We download text data for 104 languages of mBERT \cite{devlin-etal-2019-bert} from Wikipedia \eat{\cite{wikidump}}, and for 100 languages of XLM-R from the CC-100 corpus \cite{conneau-etal-2020-unsupervised}. For each language we sample 10,000 random token embeddings to compute the projection matrix for each layer, and project the multilingual representation space into a linear subspace shared by all languages, in which all languages are well separated. Thus the language-specific subspace contains interpretable information that can identify different languages.

\subsection{Probing Tasks}
After projecting the representations, we use probing tasks to analyze how much language-specific information is contained. 

In language-specific subspaces, it should be easy to determine what language the token/sentence is written in. We randomly select language pairs to train binary Logistic Regression classifiers for \textbf{Language Identification (LI)} on the subspaces and evaluate the average pairwise accuracy. That is, we train and evaluate in a one-vs-one fashion.

To investigate
\textbf{Linguistic Typology (LT)}, we use 
WALS \cite{wals}. To predict each of WALS's
192 typological features, we train a logistic regression on the subspaces in a one-vs-rest fashion, and report the micro average $F_1$ score.

Following \cite{libovicky-etal-2020-language}, we quantify the \textbf{Language Similarity (LS)} in the subspaces by V-measure. We apply hierarchical clustering to the subspaces' centroids of languages and compare the clusters with language families.

\subsection{Token Classification}
To analyze the cross-lingual transfer ability of the subspaces for downstream tasks, we use the Wikiann dataset \cite{pan-etal-2017-cross} for named entity recognition (NER) and Universal Dependencies treebanks \cite{nivrehal} for Part-of-Speech (PoS) tagging. Somehwat in between a probing task and a downstream task, we train a linear classifier with different subspace dimensions on mBERT representations to predict NER/PoS, and evaluate the zero-shot transfer performance by average $F_1$ score on all languages.
\subsection{Setup}
\paragraph{Multilingual Language Models} For mBERT and XLM-R, we use the models "bert-base-multilingual cased" and "xlm-roberta-base" from the Transformers library \cite{wolf-etal-2020-transformers}.
\paragraph{Probing Tasks} For each task, we sample 8,000 token embeddings for training and 2,000 for evaluation. We run each experiment with five different seeds and report the mean. We use Logistic Regression from the Scikit-learn library \cite{scikit-learn} and Hierarchical Clustering from the Scipy library \cite{2020SciPy-NMeth} with their default setting.
\paragraph{Token Classification} We use the script from Xtreme \cite{hu2020xtreme} and switch the dimension of the output layer and connect to the corresponding subspaces.

\section{Results}
\subsection{Probing Tasks}

\begin{figure}[t]
	\centering
	\def\wm{0.9\linewidth}
		\begin{minipage}[t]{1\linewidth}
			\centering
			\includegraphics[width=\wm]{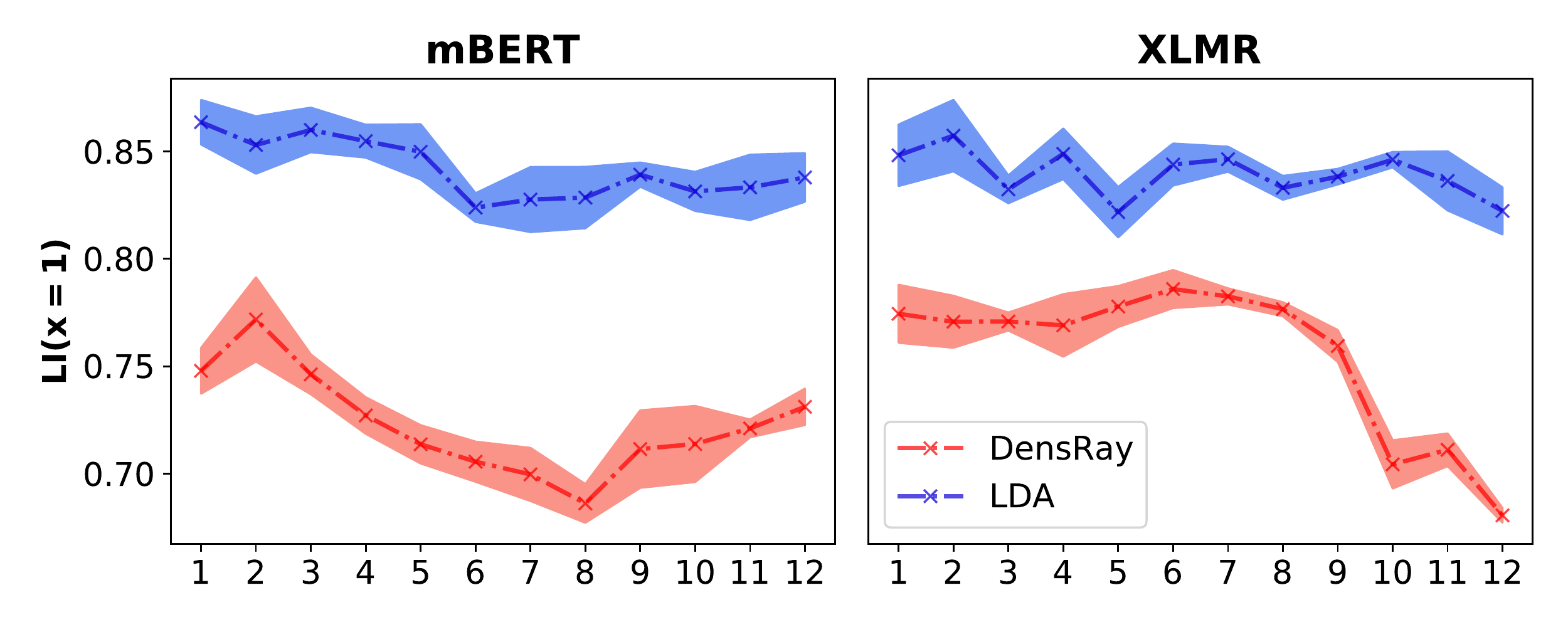}\\
			\vspace{-7pt}
			\includegraphics[width=\wm]{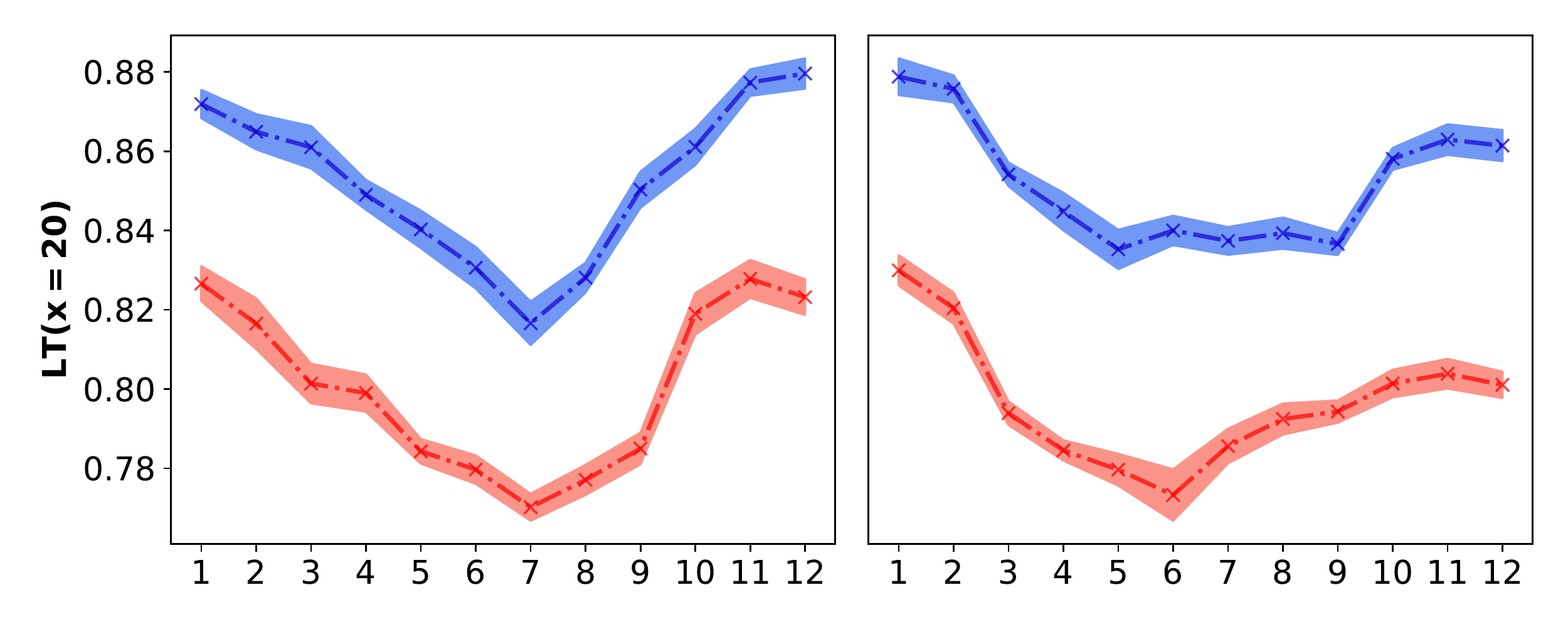}\\
			\vspace{-7pt}
			\centering
			\includegraphics[width=\wm]{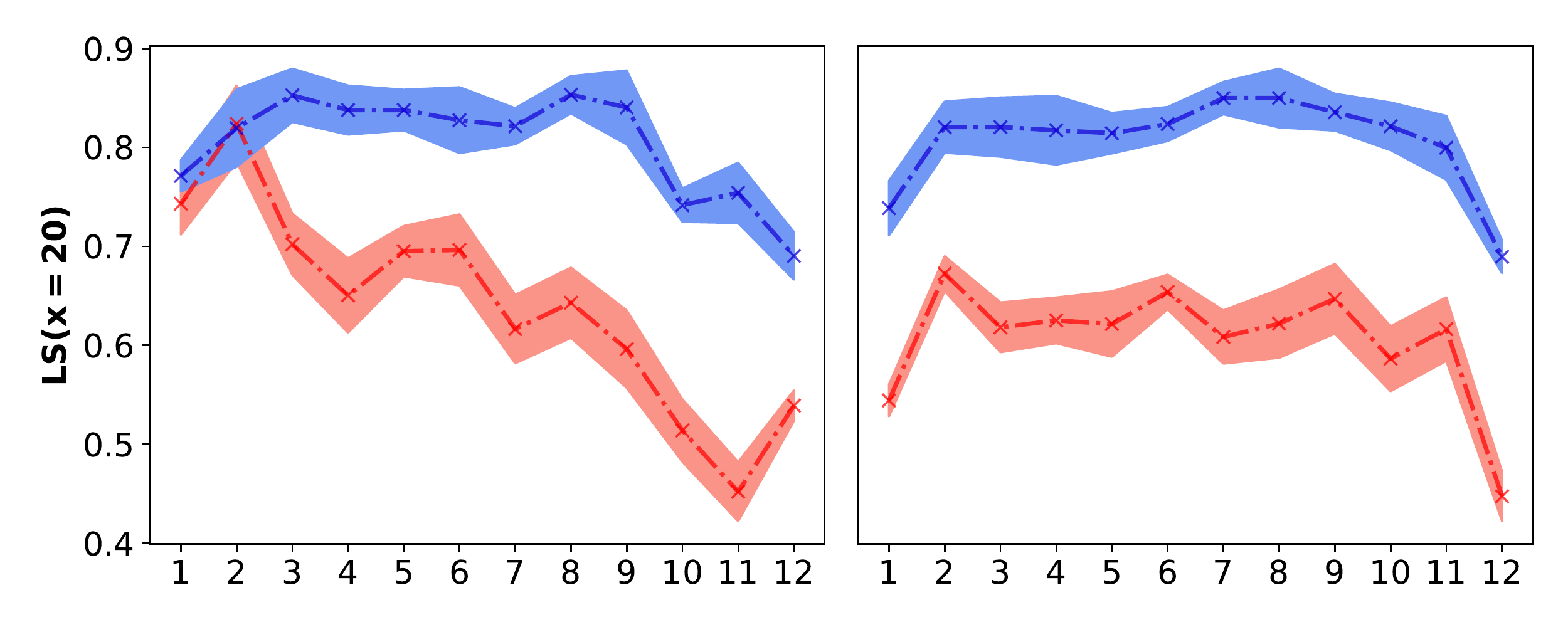}
		\end{minipage}%
	\caption{Probing tasks results  across different layers.}
	\figlabel{layers}
\end{figure}

\figref{subspace} shows the scores of the probing classifier for the different dimensions. For LI one can see nicely that the curves for using dimensions [0:$x$] consistently increase. Similarly, when using the dimensions [$x$:2$x$] the scores go up, but then decrease. This is the point where language information becomes less prevalent. The curve meets the dotted line that indicates the accuracy when choosing $x$ random dimensions at roughly $x$=100. This is an indication that up to $100$ dimensions contain all language-specific information, which matches our proof in Appendix \ref{appendix:upper}. 

For LT one can see a similar effect, yet not so pronounced. Most probably because classifying linguistic features is a more challenging task and many languages can share the same features. For LS, the solid line decreases as here we use an unsupervised clustering approach rather than training classifiers. So adding more dimensions results in more noise. 

Thus we provide a different view in \figref{diff} for the same tasks, where the main difference is, \figref{subspace} shows the dimensionality to cover \emph{all} language information is roughly equal to the number of languages, while \figref{diff} shows the dimensionality to cover \emph{sufficient} language information is different for each task.

\figref{diff} directly plots the difference between accuracy when using
dimensions [$x$-10:$x$] compared to 10 random
dimensions. We see that the dimensionality of the language specific subspace is different in
each task, e.g., in LI, language-specific subspace on the
12th layer in mBERT has dimensionality$\approx$100, while
for LT/LS, it is 70 and 30. Clearly, a one-dimensional
language-specific subspace
(as often assumed in some prior work) is not valid. 

In \figref{layers} we demonstrate the probing results with
 subspaces [0:$x$] on different layers. We select x=1 in LI while x=20 in other two tasks for better visualization, since LI is a simple task, the differences between layers will be less obvious with larger x.
 Results show that lower layers show
 strong LI and LT capabilities, which may indicate that lower layers store more language-specific information. Although the LI curves go up at the last layers, this also happens in LT while not so significant, the reason may be the learned parameters in the classifier also contribute to the results.
 Our findings are in line with prior work that finds that lower layers are more language specific \cite{muller-etal-2021-first}. Overall, the two methods, DensRay and LDA, show similar trends. \eat{We argue that the performance rise in upper layers on LI (mBERT) and LT may be caused by the classifier since this rise does not occur in LS. }

\subsection{Token Classification}
\begin{figure}[t]
	\centering
	\includegraphics[width=0.6\linewidth]{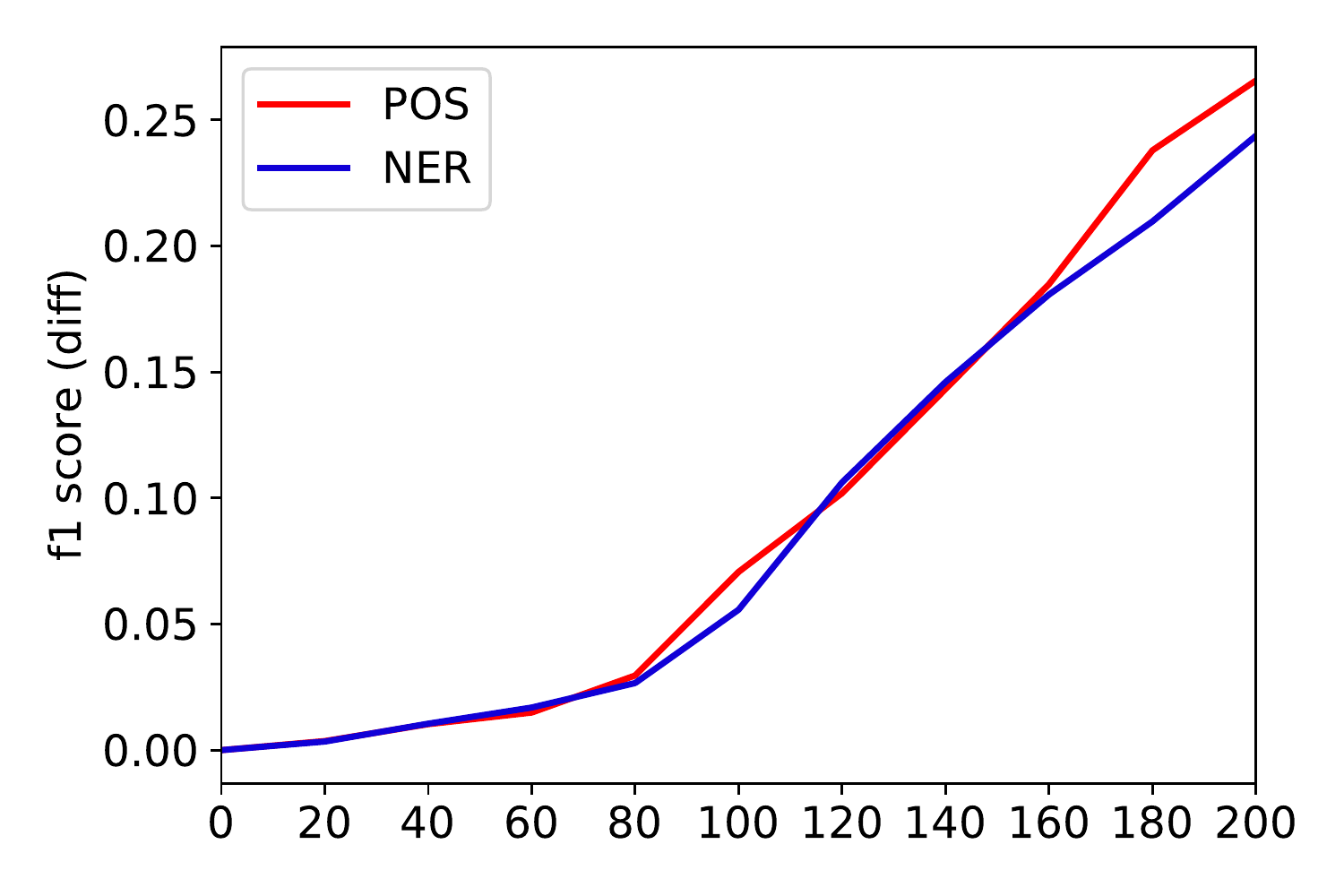}
	\caption{Token classification with subspace [$x$:768]. We report the $F_1$ difference between the classifier using [0:768] minus $F_1$ when trained on [$x$:768]. }
	\figlabel{token}
\end{figure}
 In \figref{token}
we demonstrate performance gap between using the entire space and the
subspace [$x$:768], which mainly stores language-agnostic
information for NER and PoS. We see that the curves are to zero initially, rising gradually. Omitting the first 100 dimensions only marginally affects the crosslingual performance. 
Then, the
difference grows close to linearly. This might indicate that downstream tasks would be more sensitive to the semantic information in language-agnostic subspace, and less sensitive to the language identity information in the language-specific subspace, which would insight our proposed regularization approach in Appendix \ref{appendix:table}.

\section{Conclusion}
This work follow the line of research to analyze how interpretable language information is stored in multilingual representations. We applied DensRay and LDA to locate language-specific information in linear subspaces from mBERT and XLM-R's representations. Our probing results show that language-specific information can be well decoded by the method we proposed. The dimensionality of the subspace we extracted is determined by the task, while there exists an upper bound roughly equal to the number of languages. We also found that languages are well separated in lower layers, which may indicate that lower layers store more language-specific information. 

For future work, we are interested in continuing the research for better utilizing language-specific information to achieve better zero-shot cross-lingual transfer performance.

\eat{
\section*{Acknowledgements}
This work was supported
by the European Research Council (\# 740516). 
The second author was supported by the Bavarian research institute for digital transformation (bidt) through their fellowship program.
}

\clearpage
\bibliography{anthology,custom}
\bibliographystyle{acl_natbib}

\clearpage

\appendix
\section{The Upper bound of Dimensionality}
\label{appendix:upper}
Assume that we have $n$ classes of embeddings $C_1,...,C_n$, which stand for $n$ languages in this work, $C_i\in \mathbb{R}^{l_i\times d}$. \newcite{fisher1936use} proved the rank of LDA objective limits the upper bound of subspace dimensionality to $n$-1. Follow the same idea, here we prove the dimensionality of subspace yields by DensRay has a upper bound $n$+1. 

\vspace{6pt}
With DensRay objective \cite{dufter-schutze-2019-analytical}, we get $C(n,1)=n$ pairs of languages for $L_{_=}$, and $C$($n$,2)=$n$($n$-1)/2 pairs for $L_{\neq}$. Note $c_i$ as the mean of and $C_i$, one can simply rewrite the objective into matrix form, so for each pair $(C_i,C_j)$ and $v\in C_i,w \in C_j$ we have:
\begin{small}
$$
\sum\limits_{v,w}d_{vw}d_{vw}^T = 
- l_il_j(c_ic_j^T+c_jc_i^T) 
+ l_jC_iC_i^T+l_iC_jC_j^T
$$
\end{small}

When all classes have the same number of samples $l_{i}=m$, it becomes:
\begin{small}
$$
\sum\limits_{v,w}d_{vw}d_{vw}^T =  - m^2c_ic_j^T -m^2c_jc_i^T + mC_iC_i^T+mC_jC_j^T
$$
\end{small}
Set weights $\alpha_{_=}=1/n$ and $\alpha_{\neq}=2/n(n-1)$, then for $v,w \in L_{_=}$:
\begin{small}
$$
\alpha_{_=}\sum\limits_{v,w}d_{vw}d_{vw}^T= -\frac{2m^2}{n}\sum\limits_{i=1}^nc_ic_i^T
+\frac{2m}{n}\sum\limits_{i=1}^nC_iC_i^T \notag
$$
\end{small}
And for $v,w \in L_{\neq}$, $\alpha_{\neq}\sum \limits_{v,w}d_{vw}d_{vw}^T$:
\begin{small}
\begin{equation}
\begin{aligned}
&=\frac{2m^2}{n(n-1)}[-\sum\limits_{i\neq j}(c_ic_j^T+c_jc_i^T)
+\frac{n-1}{m}\sum\limits_{i=1}^nC_iC_i^T]
\\
&=\frac{2m^2}{n(n-1)}[\sum\limits_{i=1}^nc_ic_i^T
-(\sum\limits_{i=1}^nc_i) (\sum\limits_{i=1}^nc_i)^T]
+\frac{2m}{n}\sum\limits_{i=1}^nC_iC_i^T \notag
\end{aligned}
\end{equation}
\end{small}
Let $A_1=\sum\limits_{i=1}^nc_ic_i^T$, $A_2=(\sum\limits_{i=1}^nc_i)(\sum\limits_{i=1}^nc_i)^T$, then:
$$
A =
\frac{2m^2}{n-1}(A_1-\frac{1}{n}A_2)
$$
Obviously the rank $R(A_1) \leq n$ reaches its upper bound $n$ when all the classes are completely independent, also we have $R(A_2)=1$ since $\sum\limits_{i=1}^nc_i$ is a vector. Thus:
	$$
	R(A) \leq R(A_1) + R(A_2)\leq n+1
	$$
This formula indicates that $n$+1 is the upper bound of dimensionality to cover all the information needed for DensRay. 
Specially, when $n=2$, we have $A=\frac{m^2}{2}(c_1-c_2)(c_1-c_2)^T$ and $R(A)=1$.

Also, when languages are similar, $R(A_1)$ will be less than $n$, which would lead to a smaller subspace. Here we use language family Austronesian and Italic to calculate DensRay projection matrix on mBERT respectively, for each language family we use 10 languages. As \figref{family} shows Austronesian and Italic reach zero earlier than random languages, this evidence supports our argument that DensRay yields smaller language-specific subspace when languages are similar.
\begin{figure}[t]
	\centering
	\includegraphics[width=0.7\linewidth]{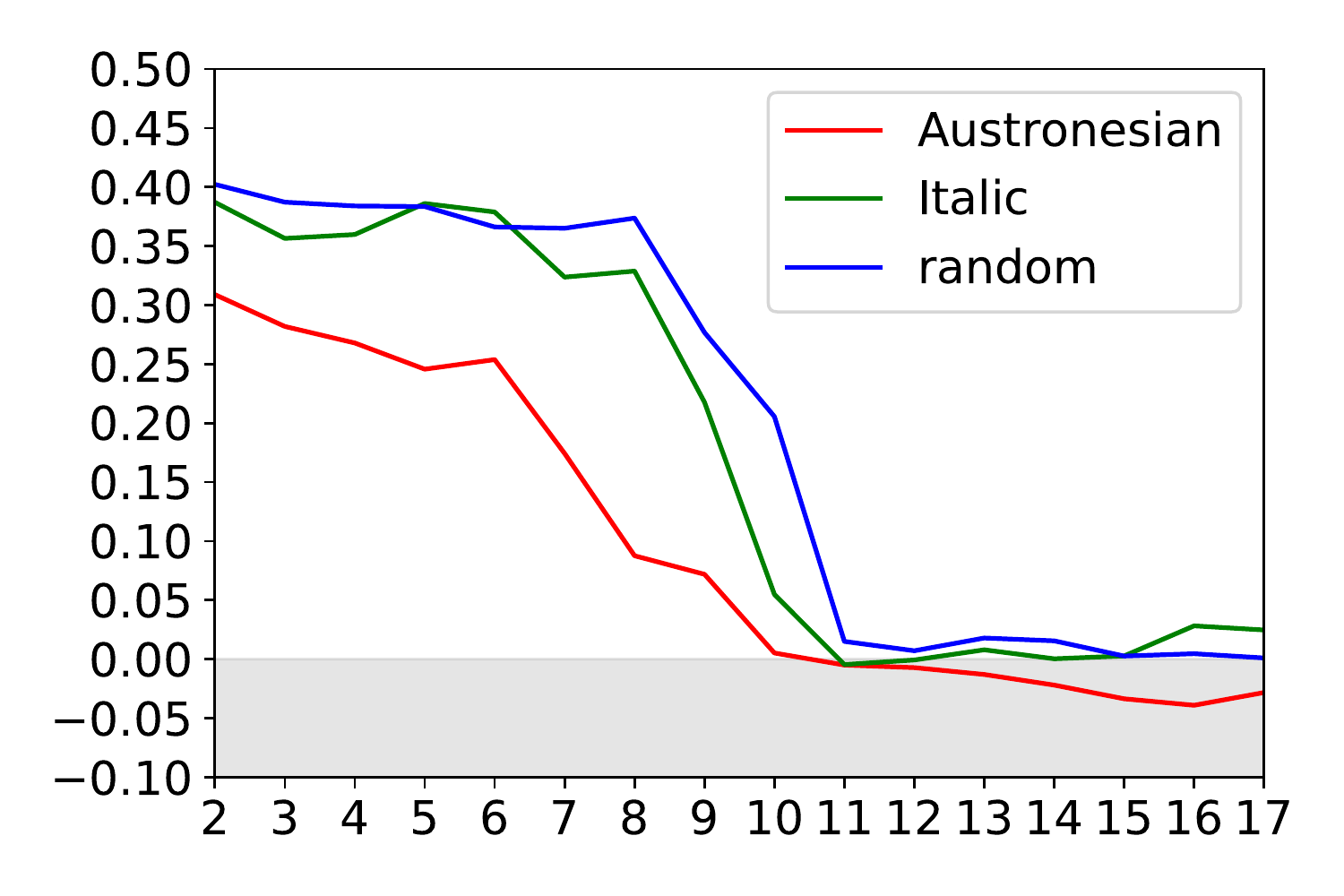}
	\caption{LI with every 2 dimensions on mBERT embeddings, projected by DesnRay, we demonstrate the performance gap with 2 random dimensions}
	\figlabel{family}
\end{figure}

\section{Activity Regularization}
\label{appendix:table}
\subsection{Methodology}
Primarily we locate the language-specific information via linear projections. However, we
attempt to use gained insights for better
zero-shot transfer and propose an activity
regularization approach. The regularizer aim to encourage the
model to keep the language-specific information close to the representations in PLM and
to change only the language-agnostic information. This
assumes that the PLM is multilingual
and finetuning on a downstream task using only a single language would destroy this multilinguality to some degree.

Let $E\in\mathbb{R}^{n\times d}$ be the original representation obtained from a pretrained multilingual encoder, $\hat{E}\in \mathbb{R}^{n\times d}$ be the representation during finetuning in a downstream task. With a DensRay transformation matrix $Q\in \mathbb{R}^{d\times d^{*}}$ where $d^{*}$ is the dimensionality of language-specific subspace, we propose L-DAR (Language subspace - DensRay Activity Regularization), a regularizer to improve the performance on zero-shot cross-lingual transfer. More specifically, we add the following regularization term to the training objective during finetuning:
$
\mathcal{L}_{DAR} = \alpha ||W(EQ-\hat{E}Q)||
$
where $\alpha$ is a weight to allow for a balance between this regularizer and the cross entropy loss, and $W\in \mathbb{R}^{d^*}$ is the explained variance ratio of the dimensions in $Q$. As DensRay is computed using an eigendecomposition we can interpret eigenvalues as explained variance. $W$ weighs individual dimensions with their contributions. \eat{to avoid hyperparameter search for dimensionality of language-specific subspace, so that in practice we set $d^{*}$ to the number of languages. }
\subsection{Setup}
We evaluate L-DAR on
the Xtreme setup  \cite{hu2020xtreme} working with
mBERT for PoS and NER, which finetuned mBERT for 2 epochs, with a training batch size of 4, a gradient accumulation step of 8, and a learning rate of 2e-5. For L-DAR, we turn off the dropout and set dimensionality $d^*$ to 105, and set $\alpha$ to 5e-3. We apply the regularizer for representations on all hidden layers. All experiments are executed on GeForce RTX 2080 Ti, which would take around 6 minutes for each epoch.

Besides using only
English training data, we also work with a multisource
transfer where we use training data from English, German and Chinese. This allows the model to learn and preserve
language differences more effectively because it sees
multiple languages during finetuning.
We evaluate zero-shot transfer with $F_1$. 

\begin{table}[t]
	\def\tablesep{0.1cm}
	\centering
	\scriptsize
	\begin{tabular}{
			@{\hspace{\tablesep}}l@{\hspace{\tablesep}}
			@{\hspace{\tablesep}}l@{\hspace{\tablesep}}|
			@{\hspace{\tablesep}}r@{\hspace{\tablesep}}
			@{\hspace{\tablesep}}r@{\hspace{\tablesep}}
			@{\hspace{\tablesep}}r@{\hspace{\tablesep}}|
			@{\hspace{\tablesep}}r@{\hspace{\tablesep}}
			@{\hspace{\tablesep}}r@{\hspace{\tablesep}}
			@{\hspace{\tablesep}}r@{\hspace{\tablesep}}
		}
		&&\multicolumn{3}{c}{PoS}&\multicolumn{3}{c}{NER}\\
		&&Train &Zero-S.&All &Train &Zero-S.&All\\
		\midrule
		\midrule
		\multirow{2}{*}{Single-Src.}&mBERT&95.3&71.0&71.8&84.1&61.2&61.8\\
		&L-DAR&\textbf{95.4}&\textbf{71.3}&\textbf{72.1}&\textbf{84.6}&\textbf{61.5}&\textbf{62.1}\\
		\midrule
		\multirow{2}{*}{Multi-Src.}&mBERT&93.8&72.1&74.1&82.5&\textbf{67.4}&\textbf{68.6}\\
		&L-DAR&\textbf{94.1}&\textbf{72.6}&\textbf{74.6}&\textbf{82.6}&66.7&68.0\\
	\end{tabular}
	\caption{Results of Zero-shot transfer. \enote{pd}{potentially add standard deviations?} \tablabel{zero-overview}}
\end{table}

\begin{table*}[t]
	\centering
	\footnotesize  
	\setlength{\abovecaptionskip}{0.1cm}   
	\setlength{\belowcaptionskip}{0cm}
	\setlength{\tabcolsep}{3pt}
	\begin{tabular}{*{18}{c}}
		\toprule
		Lang. &en & af &ar &bg &de &el  &es &et &eu &fa &fi &fr &he &hi &hu &id &it\\
		\midrule
		mBERT&
		95.3& 	86.7& 	55.7& 	84.7& 	85.5& 	81.2& 	86.7& 	79.5&		60.8& 	66.4& 	79.3& 	83.0& 	55.7& 	66.9&	78.9&	71.3&	88.4\\
		L-DAR&
		95.4&	86.0&	54.9&	86.5&	86.7&	82.6&	87.5&	80.3&
		59.3&	67.8&	79.8&	84.0&	56.2&	63.1&	79.7&	71.6&	88.2\\
		\midrule
		&ja &kk &ko &mr &nl &pt &ru &ta &te &th &tl &tr &ur &vi &yo &zh & avg\\
		\midrule
		mBERT&
		50.9& 	70.8& 	50.2& 	69.0& 	88.7& 	86.2& 	85.5& 	58.9&
		75.5& 	41.7&	82.1&	68.9&	57.8&	55.5&	58.8&	62.2&	71.8\\
		L-DAR&
		48.4&	71.1&	51.1&	74.4&	88.7&	86.7&	86.4&	60.1&
		76.7&	40.0&	83.2&	69.1&	56.8&	55.0&	57.7&	63.1&	72.1\\
		\bottomrule
	\end{tabular}
	\caption{Zero-shot (en) Transfer on POS tagging.}
	\tablabel{zeroshot_en_pos}
	\vspace{10pt}
	\centering
	\footnotesize  
	\setlength{\abovecaptionskip}{0.1cm}   
	\setlength{\belowcaptionskip}{0cm}
	\setlength{\tabcolsep}{3pt}
	\begin{tabular}{*{18}{c}}
		\toprule
		Lang.& en& de& zh& af &ar &bg &el &es &et &eu &fa &fi &fr &he &hi &hu &id\\
		\midrule
		mBERT&
		94.9&	96.2&	90.3&	87.1&	53.7&	86.7&	86.8&	86.7&	81.2&	66.3&	66.2&	80.9&	84.3&	56.8&	66.7&	80.9&	72.6\\
		L-DAR&
		95.1&	97.1&	90.2&	86.4&	54.8&	87.8&	87.1&	87.3&	80.9&
		62.0&	68.3&	81.7&	84.5&	56.7&	67.9&	81.1&	72.7\\
		\midrule
		&it &ja &kk &ko &mr &nl &pt &ru &ta &te &th &tl &tr &ur &vi &yo & avg\\
		\midrule
		mBERT&
		86.1&	46.3& 	73.0& 	52.1& 	76.3& 	90.1& 	84.0& 	86.3& 	62.9&
		75.5& 	48.8& 	82.9& 	72.1& 	57.7& 	55.4& 	57.9& 	74.1\\
		L-DAR&
		86.5&	46.5& 	73.6& 	52.5& 	76.3& 	89.6& 	84.4& 	86.7& 	62.4&
		75.4& 	49.8& 	86.7& 	72.6& 	59.7& 	56.9& 	60.5& 	74.6\\
		\bottomrule
	\end{tabular}
	\caption{Zero-shot (Multisource) Transfer on POS tagging.}
	\tablabel{zeroshot_multi_pos}
	\vspace{10pt}
	
	\centering
	\footnotesize  
	\setlength{\abovecaptionskip}{0.1cm}   
	\setlength{\belowcaptionskip}{0cm}
	\setlength{\tabcolsep}{1.5pt}
	\begin{tabular}{*{22}{c}}
		\toprule
		Lang.& en&	af&	ar&	bg&	bn&	de&	el&	es&	et&	eu&	fa&	fi&	fr&	he&	hi&	hu&	id&	it&	ja&	jv&\\
		\midrule
		mBERT&
		84.1&	76.5&	43.0&	76.7&	67.9&	77.2&	72.5&	71.2&	75.2&	64.7&	36.2&	76.8&	79.2&	57.0&	65.9&	76.0&	67.4&	80.8&	29.9&	64.7\\
		L-DAR&
		84.6&	77.3&	41.6&	77.2&	68.0&	79.1&	74.1&	77.0&	76.9&
		60.6&	38.8&	77.6&	80.2&	56.6&	65.2&	76.6&	57.7&	81.9&
		29.4&	65.3\\
		\midrule
		&ka&	kk&	ko&	ml&	mr&	ms&	my&	nl&	pt&	ru&	sw&	ta&	te&	th&	tl&	tr&	ur&	yo&	vi&	zh&	avg\\
		\midrule
		mBERT&
		64.8&	48.8&	59.7&	51.8&	56.7&	71.3&	50.2&	82.0&	79.3&
		61.7&	71.1&	50.9&	47.9&	0.64&	73.5&	73.8&	33.3&	35.1&	72.9&	43.3&	61.8\\
		L-DAR&
		65.7&	46.7&	58.3&	53.8&	57.6&	66.9&	51.2&	82.4&	81.4&	64.1&	67.1&	49.8&	47.5&	0.60&	74.5&	76.7&	33.9&	43.2&	71.8&	43.2&	62.1\\
		\bottomrule
	\end{tabular}
	\caption{Zero-shot (en) Transfer on NER.}
	\tablabel{zeroshot_en_ner}
	\vspace{10pt}
	
	\centering
	\footnotesize  
	\setlength{\abovecaptionskip}{0.1cm}   
	\setlength{\belowcaptionskip}{0cm}
	\setlength{\tabcolsep}{1.5pt}
	\begin{tabular}{*{22}{c}}
		\toprule
		Lang.& en&	de&	zh&	af&	ar&	bg&	bn&	el&	es&	et&	eu&	fa&	fi&	fr&	he&	hi&	hu&	id&	it&	ja\\
		\midrule
		mBERT&
		82.3&	87.6&	77.5&	80.8&	51.6&	82.0&	69.8&	78.4&	83.6&	81.0&	69.9&	48.5&	80.5&	85.4&	58.9&	70.0&	83.5&	65.0&	81.6&	42.7
		\\
		L-DAR&
		82.2&	87.6&	77.9&	80.6&	47.5&	82.0&	71.9&	79.2&	82.8&	80.0&	74.3&	43.4&	81.9&	84.5&	57.6&	68.3&	82.2&	71.2&	85.1&	46.1\\
		\midrule
		&	jv&	ka&	kk&	ko&	ml&	mr&	ms&	my&	nl&	pt&	ru&	sw&	ta&	te&	th&	tl&	tr&	ur&	yo&	vi&	avg\\
		\midrule
		mBERT&
		65.9&	73.1&	57.6&	67.4&	64.8&	64.6&	74.1&	55.2&	84.4&	83.3&	70.6&	64.6&	54.8&	56.6&	6.1&	74.7&	82.5&	47.0&	59.3&	75.8&	68.6\\
		L-DAR&
		65.2&	71.4&	52.6&	67.9&	62.6&	65.2&	73.5&	53.8&	85.9&	83.5&	69.0&	61.4&	57.3&	60.4&	9.1&	74.5&	78.8&	37.6&	51.9&	69.1&	68.0\\
		\bottomrule
	\end{tabular}
	\caption{Zero-shot (Multisource) Transfer on NER.}
	\tablabel{zeroshot_multi_ner}
\end{table*}
\subsection{Results}
\tabref{zero-overview} shows that with English training data, L-DAR achieves a small improvement around 0.3\% on both tasks. Thus, preserving language specific information from pretraining seems a promising research approach for future work. In multisource training with English, German and Chinese,
L-DAR outperformes the baseline with 0.5\%  on POS tagging while the average $F_1$ score drops by 0.6\% on NER task. Detailed results for each language are shown in \tabref{zeroshot_en_pos}\textasciitilde\tabref{zeroshot_multi_ner}.

\begin{table}[H]
	\def\tablesep{0.1cm}
	\centering
	\scriptsize
	\begin{tabular}{
			@{\hspace{\tablesep}}l@{\hspace{\tablesep}}
			@{\hspace{\tablesep}}l@{\hspace{\tablesep}}|
			@{\hspace{\tablesep}}r@{\hspace{\tablesep}}
			@{\hspace{\tablesep}}r@{\hspace{\tablesep}}
			@{\hspace{\tablesep}}r@{\hspace{\tablesep}}|
			@{\hspace{\tablesep}}r@{\hspace{\tablesep}}
			@{\hspace{\tablesep}}r@{\hspace{\tablesep}}
			@{\hspace{\tablesep}}r@{\hspace{\tablesep}}
		}
		&&\multicolumn{3}{c}{PoS}&\multicolumn{3}{c}{NER}\\
		&&Train &Zero-S.&All &Train &Zero-S.&All\\
		\midrule
		\midrule
		\multirow{2}{*}{Single-Src.}&mBERT&95.3&71.0&71.8&84.1&61.2&61.8\\
		&L-DAR&\textbf{95.4}&\textbf{71.3}&\textbf{72.1}&\textbf{84.6}&\textbf{61.5}&\textbf{62.1}\\
		&L-DAR $_{\triangle1-4}$&95.2&70.2&71.5&83.0&60.6&61.2\\
		&L-DAR $_{\triangle5-8}$&95.2&70.9&71.6&82.9&60.8&61.3\\
		&L-DAR $_{\triangle9-12}$&95.2&71.4&72.0&83.1&61.1&61.6\\
		\midrule
		\multirow{2}{*}{Multi-Src.}&mBERT&93.8&72.1&74.1&82.5&\textbf{67.4}&\textbf{68.6}\\
		&L-DAR&\textbf{94.1}&\textbf{72.6}&\textbf{74.6}&\textbf{82.6}&66.7&68.0\\
		&L-DAR $_{\triangle1-4}$&93.5&72.0&73.9&80.2&66.0&67.1\\
		&L-DAR $_{\triangle5-8}$&93.6&72.1&74.1&80.2&66.1&67.1\\
		&L-DAR $_{\triangle9-12}$&93.7&72.3&74.2&80.2&66.4&67.4
	\end{tabular}
	\caption{Layer Analysis for Zero-shot transfer. \tablabel{zero-layers}}
\end{table}
\subsection{Analysis}
As an additional analysis we apply the regularizer for representations from every 4 layers e.g. $\triangle 1-4$. 
Here we set $\alpha$ to 2e-2. 
\tabref{zero-layers} shows that L-DAR achieves better performance on upper layers, especially, on both single source task and multisource PoS, L-DAR shows comparable $F_1$ with the baseline. However this behaviour is not shown on multilingual NER, which leads to the drop in this task, we left this problem for further research.

\end{document}